\documentclass[sigconf]{acmart}
\makeatletter
\renewcommand\@formatdoi[1]{\ignorespaces}
\makeatother
\renewcommand\footnotetextcopyrightpermission[1]{} 
\AtBeginDocument{%
  \providecommand\BibTeX{{%
    \normalfont B\kern-0.5em{\scshape i\kern-0.25em b}\kern-0.8em\TeX}}}

\usepackage{indentfirst}
\usepackage{multirow}
\usepackage{rotating}
\usepackage{graphicx}
\usepackage{caption}
\usepackage{subcaption}
\usepackage{fancyhdr}
\begin{document}

\title{Importance Weighted Adversarial Discriminative Transfer for Anomaly Detection}
\author{Cangning Fan}
\authornote{This work is completed during the internship at Alibaba DAMO Academy.}
\email{fancangning@gmail.com}
\author{Fangyi Zhang, Peng Liu, Xiuyu Sun, Hao Li, Ting Xiao, Wei Zhao, Xianglong Tang}

\renewcommand{\shortauthors}{Cangning Fan, et al.}

\begin{abstract}
    Previous transfer methods for anomaly detection generally assume the availability of labeled data in source or target domains. However, such an assumption is not valid in most real applications where large-scale labeled data are too expensive. Therefore, this paper proposes an importance weighted adversarial autoencoder-based method to transfer anomaly detection knowledge in an unsupervised manner, particularly for a rarely studied scenario where a target domain has no labeled normal/abnormal data while only normal data from a related source domain exist. Specifically, the method learns to align the distributions of normal data in both source and target domains, but leave the distribution of abnormal data in the target domain unchanged. In this way, an obvious gap can be produced between the distributions of normal and abnormal data in the target domain, therefore enabling the anomaly detection in the domain. Extensive experiments on digit image datasets and the UCSD benchmark demonstrate the effectiveness of our approach.
\end{abstract}



\maketitle
\thispagestyle{empty}

\section{Introduction}
    Anomaly detection is a fundamental data analysis task. The goal of anomaly detection is to identify anomalous instances called anomalies which do not conform to the expected normal pattern \cite{chandola2009anomaly}. Anomaly detection has been used widely in practice as anomalies often correspond to substantial problems that could cause significant losses, such as fraud detection \cite{kou2004survey}, medical care \cite{keller2012hics}, and intrusion detection \cite{dokas2002data}.
    
    It is called {\itshape unsupervised anomaly detection} \cite{breunig2000lof,ramaswamy2000efficient} when there is no available labeled normal/abnormal data in the dataset. Generally, unsupervised methods make use of the assumption that anomalies rarely occur, which means that they fall in a low-density area of the instance space or are far away from normal instances. Such as system maintenance this kind of situation rarely occurs, in fact, it is not an abnormal event. This severely deteriorates the performance of unsupervised anomaly detection. In some cases, the label that indicates a normal/abnormal instance can be used. Labeled data can correct errors caused by unsupervised anomaly detectors. By using anomaly labels, {\itshape fully supervised anomaly detection} \cite{chawla2002smote,herschtal2004optimising} approaches can detect anomalies much better than unsupervised ones. However, since it is usually difficult to obtain anomaly labels, because anomalies rarely occur, a fully supervised approach is usually not feasible. In other cases, only normal instances can be used to obtain anomaly detectors. The {\itshape semi-supervised anomaly detection} \cite{memae,deeponeclass} approaches are more feasible than fully supervised ones since normal instances are relatively easy to prepare, and are better than unsupervised ones since semi-supervised learning improves the accuracy of identifying anomalies.
    
    People usually expect that an anomaly detection system in the real world can monitor many scenes, each of which may be only slightly different. This can happen when monitoring products in assembly lines, illegal behavior on the street, or crops on a farm. These anomaly detection tasks are similar to each other, which stimulates people's interest in using transfer learning to transfer labeled instances from one task to another. Transfer knowledge between anomaly detection tasks which is called anomaly detection transfer have been proposed to learn anomaly detection by using normal and abnormal instances in the source domain. These methods also use target normal instances for training \cite{andrews2016transfer,latentdomainrepresentation,LocIT,ide2017multi,xiao2015robust}. However, training with abnormal labels in some applications may cause problems. There is rarely enough time and resources to mark instances in the source and target domains. Since anomalies rarely occur and labeling instances are very expensive, it is more likely that there are only normal instances in the source domain and only unlabeled normal/abnormal instances in the target domain.
    
    In this paper, we propose a novel method to improve the anomaly detection performance on target domain with no labeled instances by using only normal instances in source domain. Our method consists of an autoencoder, a domain classifier and an importance weighted module. The autoencoder tries its best to reconstruct the input. The domain classifier recognizes the difference between the source domain instance distribution and the target domain instance distribution to find the stricter lower bound of the true domain distribution divergence, while the encoder assigns the distribution divergence by stepping in the opposite direction to the domain classifier. The importance weighted module uses the reconstruction loss of the sample to weight the process of adversarial learning and reconstruction, so that the normal samples of the target domain play a greater role in the training process. Specifically, the normalized output of the inverse proportional function of the reconstruction loss gives the probability that the instance comes from the normal instances. The intuition of the weighting scheme is that if the normalized output is large, the autoencoder can reconstruct the samples almost perfectly. Therefore, a larger weight is assigned to the instance, the sample is likely to come from a normal sample in the target domain. Thus, we use the reconstruction loss as an indicator of the importance of each instance in the target domain. Then we apply the learned weight to the target sample, and input the weighted target samples into the domain classifier and autoencoder for optimization in the next iteration. Referring to \cite{0017DLO18}, we have proved that the minimax two-player game between the autoencoder and the domain classifier is theoretically equivalent to reducing the Jensen-Shannon divergence between the weighted target density and the source density.
    \begin{figure*}
        \centering
        \includegraphics[width=\textwidth]{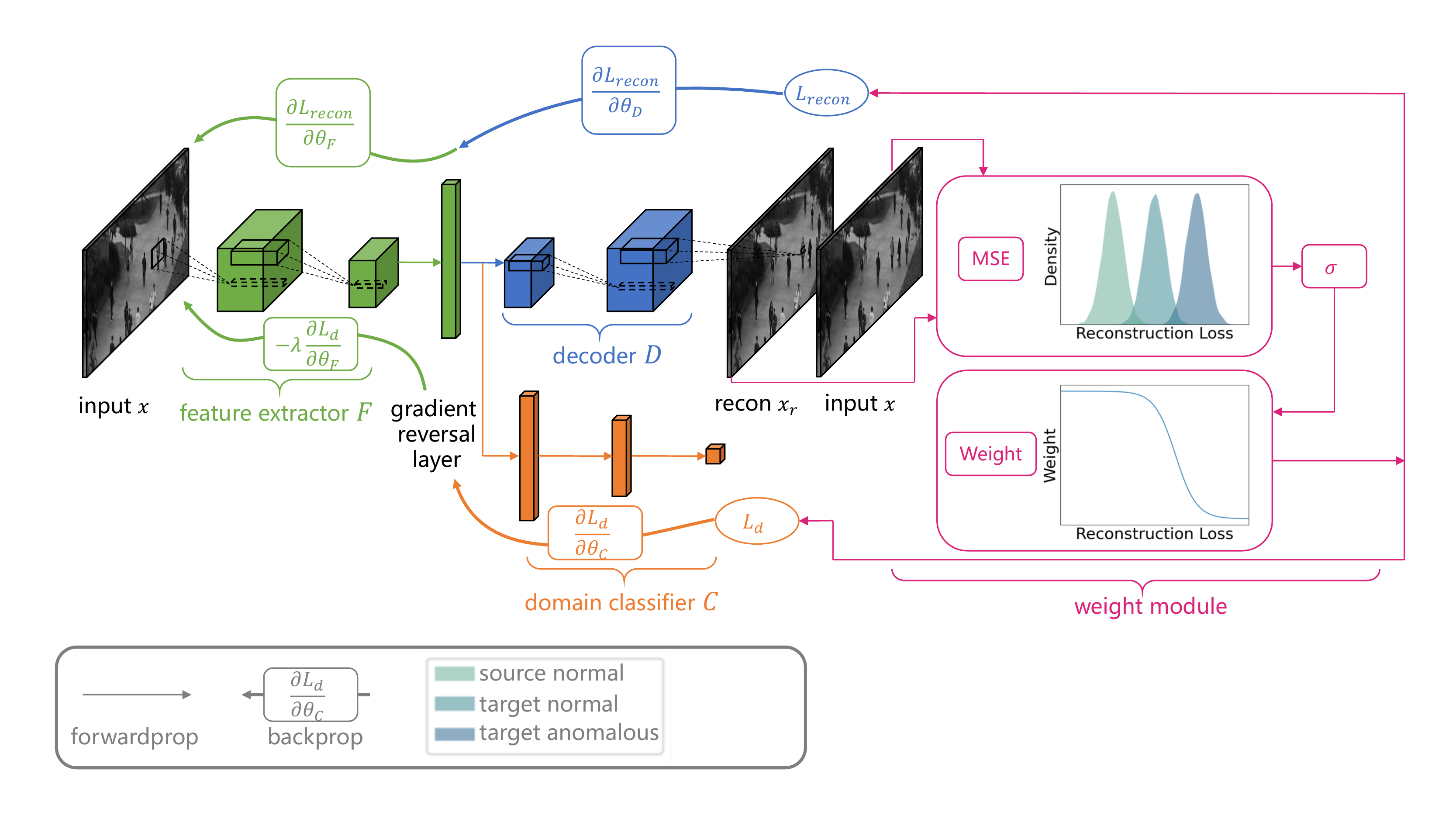}
        \caption{The overview of the proposed method. The green parts are the feature extractor for source and target domain. The blue parts are the decoder for reconstructing the input $x$. The orange parts are the domain classifier that plays the minimax game with the weighted target domain samples and the source samples. The pink parts are the weight module which weight the samples according to the reconstruction loss in order to make the source normal and target normal instances play a major role in reconstruction and adversarial learning. The gradient reversal layer is an identity transformation in the forward propagation while changes the sign of the gradient in the backward propagation.}
        \label{fig:network}
    \end{figure*}
    
\section{Related Work}
    In this section, we discuss the most related work in anomaly detection, transfer learning and domain adaptation transfer.
    
    \textbf{Anomaly Detection.} Anomaly detection, which is also called outlier detection or novelty detection, has been widely studied\cite{chandola2009anomaly,chalapathy2019deep}. Semi-supervised anomaly detection is most studied in this field. Many approaches have been proposed such as AE based methods\cite{memae,akcay2018ganomaly,park2020learning}, OSVM based methods\cite{chalapathy2018anomaly,deeponeclass}, and density based methods\cite{zhai2016deep,zong2018deep}. In addition, unsupervised anomaly detection assumes that training set contains some anomalies. Three most popular kinds of unsupervised approaches are local density-based methods\cite{breunig2000lof,papadimitriou2003loci}, k-nearest neighbor detectors\cite{ramaswamy2000efficient}, and isolation methods\cite{liu2008isolation}. Both semi-supervised and unsupervised approaches cannot use anomaly labels, leading to a limited performance. Fully supervised anomaly detection\cite{chawla2009data,haixiang2017learning} makes use of anomaly labels to improve the performance. However, all the three kinds of anomaly detection paradigms assume that all instances are sampled from the same distribution and thus cannot perform well in another domain since there is a distribution shift between source and target domain. 
    
    \textbf{Transfer Learning.} The goal of transfer learning or domain adaptation is to perform a task in target domain by using knowledge obtained from source domain. To transferring knowledge, one class of methods, such as DAN \cite{dan}, JAN \cite{jan}, RTN \cite{rtn}, align the source and target domains by reducing the maximum mean discrepancy. A second class of methods, such as ADDA \cite{adda}, MADA \cite{mada} and CDAN \cite{cdan}, attempt to apply adversarial learning to the training of network. The differences between domains are gradually eliminated since the feature extractor keeps trying confusing the domain classifier. Finally, methods such as GTA \cite{GTA} and CycleGAN \cite{CycleGAN} generate fake labeled target instances to complete target tasks by transferring the style of labeled source instance to target domain. All these methods above do not assume the class-imbalance and thus are suitable for anomaly detection. Although there are several methods\cite{al2016transfer,ge2014handling} specifically designed for the class-imbalance, they assume anomalous instances in the target domain, which is not necessary in our proposed method.
    
    \textbf{Transfer Domain Adaptation.} There are only a handful of papers exploring transfer learning for anomaly detection. Atsutoshi et al. \cite{kumagai2019transfer} infers the anomaly detectors for target domains without re-training by introducing the concept of latent domain vectors. Andrews et al. \cite{andrews2016transfer} tries to reuse learned image representations across different image datasets. Xiao et al. \cite{xiao2015robust} designs a robust one-class transfer learning method. Unlike our proposed method, the later three approaches require labeled target instances to construct the detector. The lack of labels in anomaly detection task that motivated our approach invalidates these approaches. LOCIT \cite{vincent2020transfer} selects labeled source instances to transfer by a local distribution-based approach and constructs a KNN classifier based on these selected source instances and unlabeled target instances. Although LOCIT \cite{vincent2020transfer} has the ability to handle the situation where source domain only contains normal instances, this method degenerates into a KNN-based unsupervised anomaly detection method without knowledge transfer.

\section{Proposed Method}
    \subsection{Definition}
        In order to better describe the problem, in this section, we first give the definition of terminology. We denote the source samples and the target samples as $X_s \in \mathbb{R}^{D \times n_s}$ and $X_t \in \mathbb{R}^{D \times n_t}$ respectively. The source samples and the target samples obey the distribution $p_s(x)$, $p_t(x)$ respectively. $D$ is the dimension of samples, $n_s$, $n_t$ respectively indicate the number of samples in the source domain and the number of samples in the target domain. Our method is proposed to solve the anomaly detection transfer problem setting where there are only normal instances $\mathcal{D}_{s}=\{x_i^{s}\}_{i=1}^{n_s}, x_i^{s} \in \mathbb{R}^D$ in source domain and unlabeled normal/anomalous instances $\mathcal{D}_{t}=\{x_j^{t}\}_{j=1}^{n_t}, x_j^{t} \in \mathbb{R}^D$ in target domain. The source domain and the target domain share the same feature space: $\mathcal{X}_s = \mathcal{X}_t$. The training process is divided into two stages: pretrained stage which is used to construct the suitable reconstruction loss distribution for the weighted module, and adversarial learning which is used to align the distributions of source samples and target normal samples.
    \subsection{Pretrained Stage}
        The purpose of the pretrained stage is to ensure that the source reconstruction loss is the lowest, followed by the target normal sample reconstruction loss, and the target abnormal sample reconstruction loss is the highest. Under this distribution, weighted adversarial learning is used to align the source distribution and the target normal distribution to make the reconstruction loss of the target normal sample smaller and smaller, and the reconstruction loss of the target abnormal sample remains unchanged. In this way, the difference between the reconstruction loss of the normal sample and the abnormal sample is significantly increased, and the anomaly detection is realized. We propose a unbalanced training procedure to train AE for reconstruction tasks.
        \begin{equation}
            \begin{aligned}
                \min_{F,D}\mathcal{L}_{recon} = \mathbb{E}_{x\sim p_s(x)}L(D(F(x)), x) \\+ \lambda\mathbb{E}_{x\sim p_t(x)}L(D(F(x)), x)
            \end{aligned}
        \end{equation}
        where $L$ is the empirical loss for reconstruction task and the mean squared error loss is used in our paper. $\lambda$, which ranges from 0 to 1, is a hyperparameter used to balance the role of target samples in $\mathcal{L}_{recon}$.
        
    \subsection{Adversarial Learning Stage}
        ~\\
        \subsubsection{Adversarial learning}
        ~\\
        
        Adversarial learning is a common and effective technology in domain adaptation to align the distributions of the source domain and target domain\cite{adda, cdan}. Its basic idea is to build a competitive relationship between the feature extractor $F$ and the domain discriminator $C$: the domain discriminator $C$ tries to judge that the feature come from the source domain or the target domain, and the feature extractor $F$ tries to confuse the domain discriminator $C$. In this way, the source domain distribution and the target domain distribution are aligned. The idea of adversarial learning comes from GAN\cite{GAN}, and the definition of adversarial loss is as follows:
        \begin{equation}
            \begin{aligned}
                \min_{F_s,F_t}\max_C \mathcal{L}(C,F_s,F_t)=\mathbb{E}_{x\sim p_t(x)}[log C(F_t(x))]\\+\mathbb{E}_{x\sim p_s(x)}[log(1-C(F_s(x)))]
            \end{aligned}
        \end{equation}
        where $F_s$ and $F_t$ are the feature extractors , and $C$ is the domain classifier which classifies the source samples as 0 and classifies the target samples as 1. In this paper, the source domain samples and the target domain samples share the feature extractor $F$, and the counter loss is changed to:
        \begin{equation}
            \label{equation:adversarialloss}
            \begin{aligned}
                \min_{F}\max_{C}\mathcal{L}(C, F) = \mathbb{E}_{x\sim p_t(x)}[log C(F(x))]\\+\mathbb{E}_{x\sim p_s(x)}[log(1-C(F(x)))]
            \end{aligned}
        \end{equation}
        
        \subsubsection{Importance weighted adversarial learning}
            After the pretraining stage, the reconstruction loss of each sample can be obtained by the following formula:
            \begin{equation}
                \begin{aligned}
                        \mathcal{L}_{recon}(x) = L_{mse}(D(F(x)),x)
                \end{aligned}
            \end{equation}
            where $L_{mse}$ is the mean squared error loss. Suppose that the AE has converged to its optimal parameters, the $\mathcal{L}_{recon}(x)$ can be used to indicate which distribution(source normal, target normal or target anomalous) the instance $x$ comes from. Specifically, if the $\mathcal{L}_{recon}(x)\approx 0$, then the sample is highly likely to come from source normal instances, since the source normal instances play the most important role in the reconstruction loss in the pretraining stage and thus can be reconstructed best by AE. On the other hand, if $\mathcal{L}_{recon}(x)$ is not too small, the instance is more likely from target normal instances. These samples should be also given a larger importance weight to reduce the domain shift on the source and target normal instances. Finally, if $\mathcal{L}_{recon}(x)$ is large, the instance is more likely from target anomalous instances, since target anomalous instances account for only a small proportion of the target instances and thus AE can not learn these instances adequately. Hence, the weight function should be inversely related to $\mathcal{L}_{recon}(x)$ and a natural way to define the importance weights function of the instances is:
            \begin{equation}
                \begin{aligned}
                        \widetilde{w}(x) = \sigma(\eta\mathcal{L}_{recon}(x)+\beta)
                \end{aligned}
            \end{equation}
            where $\sigma$ is the sigmoid function, $\eta$ and $\beta$ are the hyperparameters which control the shape and position of $\sigma$ function respectively($\eta<0,\beta>0$). It can be seen that if $\mathcal{L}_{recon}(x)$ is large, $\widetilde{w}(x)$ is small. Hence, the weights for target anomalous instances will be smaller than the normal instances.
        
            After the introduction of the importance weighting module, the training objective to adversarial loss is changed to:
            \begin{equation}
                \label{equation:minimaxgame}
                \begin{aligned}
                        \min_F\max_C\mathcal{L}_w(C, F)=\mathbb{E}_{x\sim p_t(x)}[w(x)log C(F(x))]\\+\mathbb{E}_{x\sim p_s(x)}[log(1-C(F(x)))]
                \end{aligned}
            \end{equation}
    
        \subsubsection{Overall objective} 
        ~\\
        In summary, the overall optimization goal in the pretraining phase is:
        \begin{equation}
            \begin{aligned}
                \min_{F,D}\mathcal{L}_{recon} = \mathbb{E}_{x\sim p_s(x)}L(D(F(x)), x) \\+ \lambda\mathbb{E}_{x\sim p_t(x)}L(D(F(x)), x)
            \end{aligned}
        \end{equation}
        
        The overall optimization goal in the weighted adversarial learning phase is:
        \begin{equation}
            \begin{aligned}
                    \min_{F,D}\mathcal{L}_{recon}&=\mathbb{E}_{x\sim p_s(x)}L(D(F(x)),x)\\&+\lambda\mathbb{E}_{x\sim p_t(x)}w(x)L(D(F(X)),x)
                    \\\min_F\max_C\mathcal{L}_w(C,F)&=\mathbb{E}_{x\sim p_t(x)}[w(x)log C(F(x))]\\&+\mathbb{E}_{x\sim p_s(x)}[log(1-C(F(x)))]
            \end{aligned}
        \end{equation}
        The optimization process can be divided into two stages. In the first stage, $F$ and $D$ are pre-trained by $\mathcal{L}_{recon}$ in order to obtain a suitable reconstruction loss distribution. In the second stage, the $F$, $D$, $C$ are optimized simultaneously by weighted reconstruction loss and weighted adversarial loss. The proposed architecture can be found in Figure \ref{fig:network}.
        \begin{figure*}
            \begin{subfigure}{.33\textwidth}
              \centering
              \includegraphics[width=\linewidth]{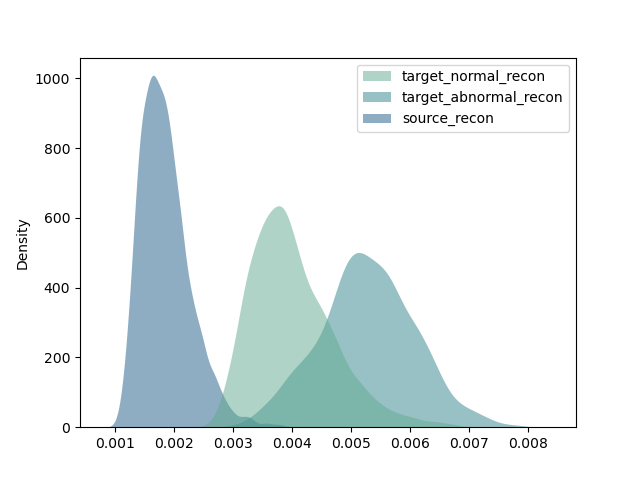}  
              \caption{0 epoch}
            \end{subfigure}
            \begin{subfigure}{.33\textwidth}
              \centering
              \includegraphics[width=\linewidth]{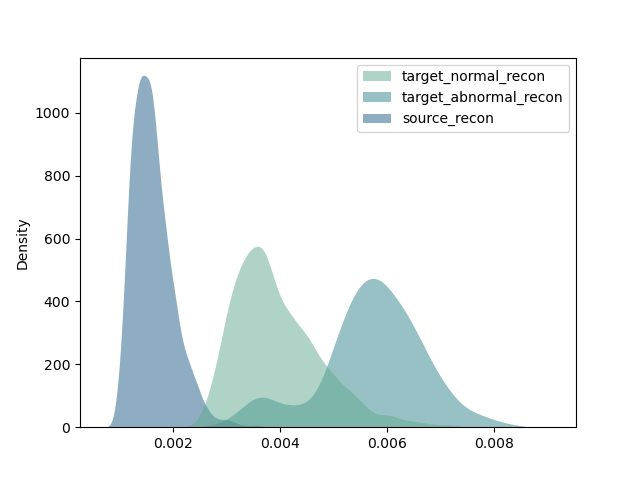}  
              \caption{19 epoch}
            \end{subfigure}
            \begin{subfigure}{.33\textwidth}
              \centering
              \includegraphics[width=\linewidth]{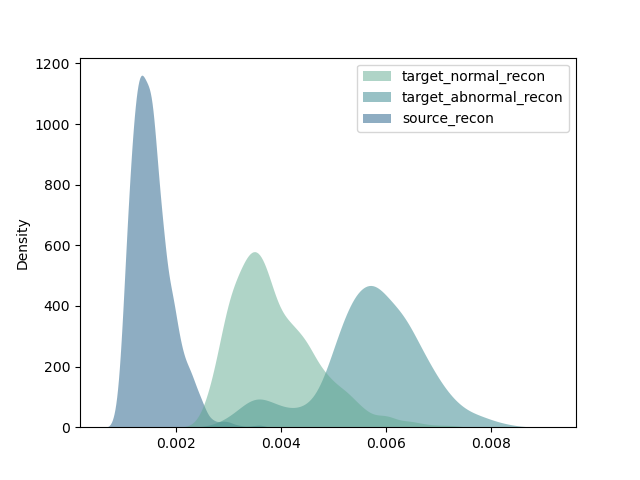}  
              \caption{29 epoch}
            \end{subfigure}
            \begin{subfigure}{.33\textwidth}
              \centering
              \includegraphics[width=\linewidth]{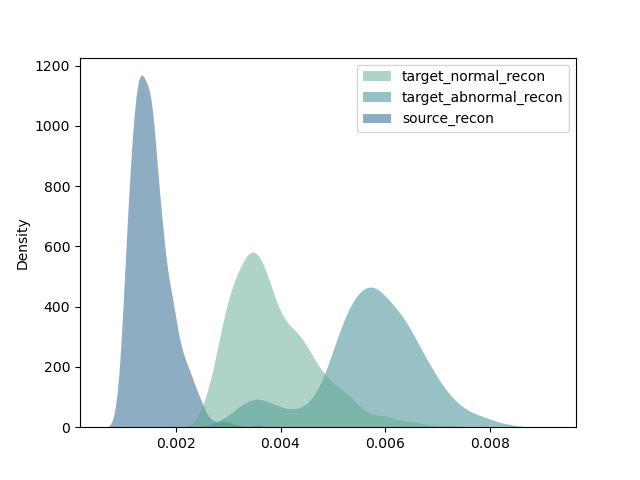}  
              \caption{39 epoch}
            \end{subfigure}
            \begin{subfigure}{.33\textwidth}
              \centering
              \includegraphics[width=\linewidth]{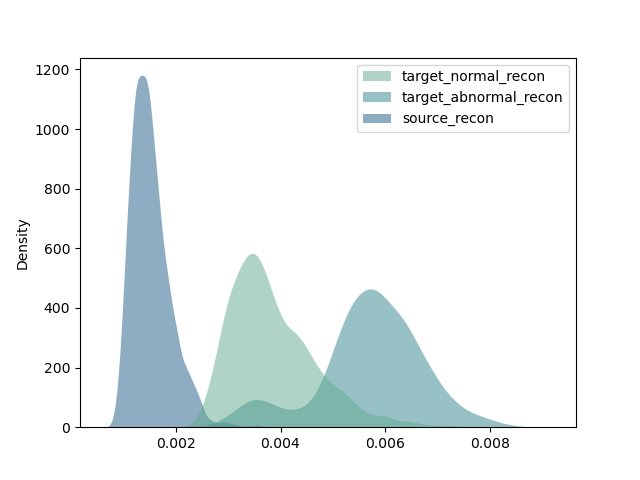}  
              \caption{49 epoch}
            \end{subfigure}
            \begin{subfigure}{.33\textwidth}
              \centering
              \includegraphics[width=\linewidth]{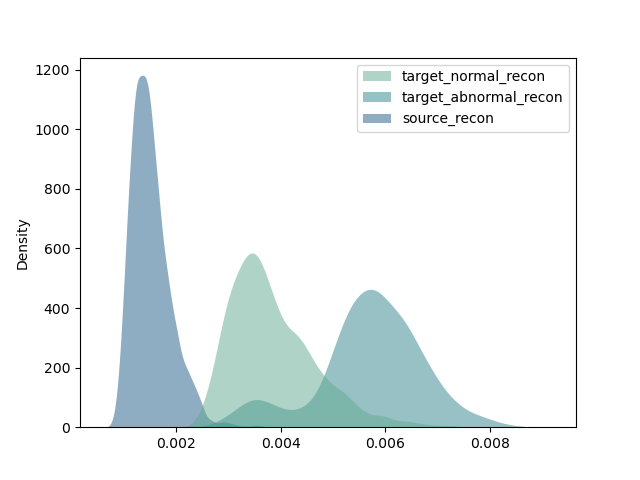}  
              \caption{59 epoch}
            \end{subfigure}
            \begin{subfigure}{.33\textwidth}
              \centering
              \includegraphics[width=\linewidth]{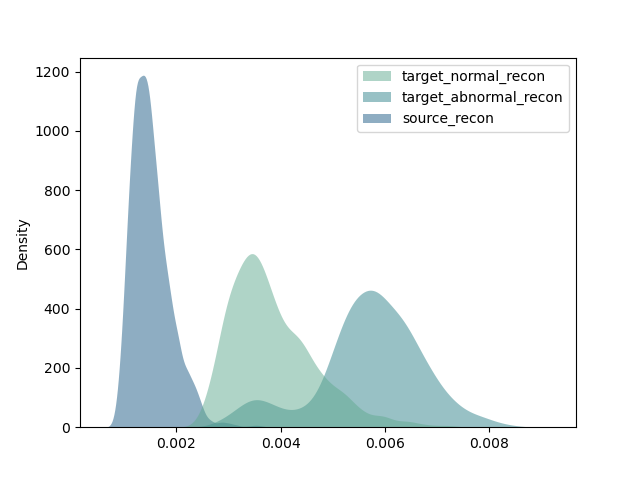}  
              \caption{69 epoch}
            \end{subfigure}
            \begin{subfigure}{.33\textwidth}
              \centering
              \includegraphics[width=\linewidth]{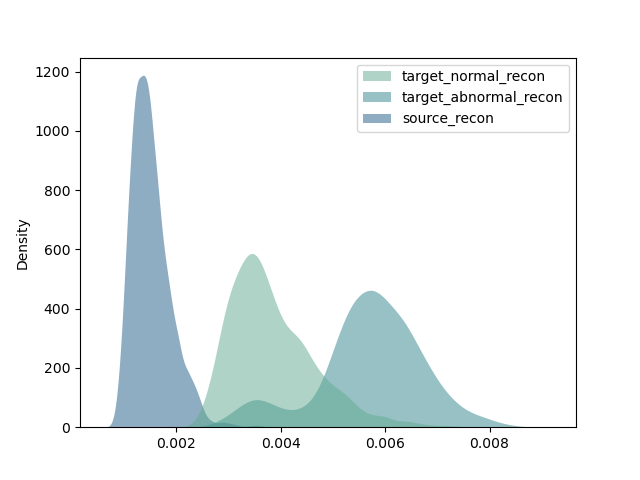}  
              \caption{79 epoch}
            \end{subfigure}
            \caption{The reconstruction loss distribution of source samples and target samples in different epoch. As training progresses, weighted adversarial learning gradually aligns the distribution of normal samples in source and target domain, at the same time, ensuring that the distribution of target anomalies remains unchanged. The reconstruction loss distribution of normal samples and abnormal samples in target domain is gradually pulled apart.}
            \label{fig:recondistribution}
            \end{figure*}
        \begin{figure}
            \begin{subfigure}{.3\textwidth}
              \centering
              \includegraphics[width=\linewidth]{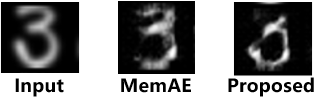}  
              \caption{Testing on the abnormal '3'}
            \end{subfigure}
            \begin{subfigure}{.3\textwidth}
              \centering
              \includegraphics[width=\linewidth]{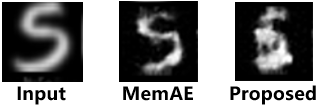}  
              \caption{Testing on the abnormal '5'}
            \end{subfigure}
            
            \begin{subfigure}{.3\textwidth}
              \centering
              \includegraphics[width=\linewidth]{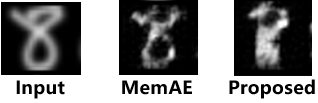}  
              \caption{Testing on the abnormal '8'}
            \end{subfigure}
            \begin{subfigure}{.3\textwidth}
              \centering
              \includegraphics[width=\linewidth]{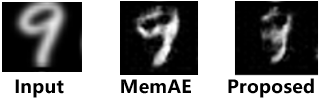}  
              \caption{Testing on the abnormal '9'}
            \end{subfigure}
            \caption{Visualization of the reconstruction results of MemAE(fine-tune) and our method on USPS. (a) The input is an image of '3'. (b) The input is an image of '5' (c) The input is an image of '8' (d) The input is an image of '9'}
            \label{fig:recon}
            \end{figure}
        
\section{Experiments}
    In this section, we verify the proposed anomaly detection transfer method. In order to prove the versatility and applicability of the proposed method, we conduct experiments on image dataset and video dataset respectively. We compare the results with different baseline models and the state-of-the-art unsupervised anomaly detection technology. We also conducted a qualitative analysis of the method through some experiments to reveal the working principle of our method. Our experimental code is implemented using PyTorch \cite{paszke2017automatic}, the network is optimized using the optimizer Adam \cite{kingma2014adam}, and the learning rate is set to 0.01.
    
    In this section, we test the performance of our proposed method through multiple experiments. In order to demonstrate the generality of our method, we not only conducted experiments on image datasets, but also conducted experiments on video datasets. The experiment is divided into two parts. The first part is the comparative experiment, which shows the effectiveness of our method by comparing the performance with the baseline methods. The second part is qualitative analysis. This part of the experimental results are used to show how our method works. The experimental code is completed using pytorch \cite{paszke2017automatic}, and the code is available at this address. The code is available at https://github.com/fancangning/anomaly\_\\detection\_transfer
    
        \begin{table*}[ht]
        \centering
        \caption{Experimental results on digit dataset. Average AUC values on two transfer tasks(Mnist$\rightarrow$USPS and USPS$\rightarrow$MNIST).}
        \begin{tabular}{c|c|ccccc}
        \hline
        \multicolumn{1}{l|}{\textbf{Task}} & \multicolumn{1}{l|}{\textbf{Anomaly rate}} & \multicolumn{1}{l}{\textbf{IsolationForest}} & \multicolumn{1}{l}{\textbf{MemAE(directly)}} & \multicolumn{1}{l}{\textbf{MemAE(fine-tune)}} & \multicolumn{1}{l}{\textbf{Our Method}} & \multicolumn{1}{l}{\textbf{MemAE}} \\ \hline
        \multirow{5}{*}{M$\rightarrow$U}   & 0.05                                       & \textbf{94.0}                                & 32.0                                         & 80.7                                          & 84.0                                    & 86.7                               \\
                                           & 0.15                                       & \textbf{97.0}                                & 34.0                                         & 68.7                                          & 83.3                                    & 88.0                               \\
                                           & 0.25                                       & 81.7                                         & 32.0                                         & 67.3                                          & \textbf{82.0}                           & 88.0                               \\
                                           & 0.35                                       & 77.7                                         & 32.3                                         & 65.3                                          & \textbf{81.7}                           & 86.7                               \\
                                           & 0.45                                       & 73.0                                         & 34.3                                         & 61.3                                          & \textbf{82.0}                           & 88.3                               \\ \hline
        \multirow{5}{*}{U$\rightarrow$M}   & 0.05                                       & 95.7                                         & 59.7                                         & 97.0                                          & \textbf{96.0}                           & 100.0                              \\
                                           & 0.15                                       & 91.7                                         & 59.5                                         & 88.0                                          & \textbf{93.0}                           & 99.7                               \\
                                           & 0.25                                       & 86.7                                         & 59.7                                         & 84.7                                          & \textbf{92.7}                           & 99.7                               \\
                                           & 0.35                                       & 82.7                                         & 60.0                                         & 79.3                                          & \textbf{91.3}                           & 99.7                               \\
                                           & 0.45                                       & 78.3                                         & 57.7                                         & 75.3                                          & \textbf{89.0}                           & 99.3                               \\ \hline
        \end{tabular}
        \label{tab:digit}
        \end{table*}
    
    \subsection{Comparative Experiments}
        We firstly conduct experiments on image dataset to compare the performance of the method with other baselines in detecting image anomalies. The two common image datasets: MNIST\cite {MNIST} and USPS are used in this experiment. These two datasets are both handwritten digital image datasets which contain the digit categories 0 to 9. For each dataset, the anomaly detection task is constructed as follows: sampling images from class 0 as normal samples, and then sampling images from other classes as abnormal samples. Our method is focus on the rarely studied anomaly detection transfer scenario, that is, the target domain has no labeled normal/abnormal data, but only normal data from the source domain exists. Therefore, for the source domain data, we only sample from category 0 to construct a normal sample set, while, for the target domain data, the dataset is composed of both the normal samples and the abnormal samples. It is worth noting that there is no normal/abnormal label in target domain. We set the abnormal sample ratio to 0.05 to 0.45 in each dataset in order to test the performance of our method on different abnormal sample ratios.
        
        In this experiment, we use the convolutional neural network to compose the encoder $F$ and decoder $D$. Refer to \cite{memae}, Conv2($k$, $s$, $c$) and Dconv2($k$, $s$, $c$) are defined to represent a 2D convolutional layer and a 2D deconvolution layer respectively, where $k$, $s$ and $c$ represent the kernel size, stride size and the number of channels. Following the settings used in \cite{memae}, we use three convolutional layers to implement the encoder: Conv2(3,2,32)-Conv2(3,2,16)-conv2(3,3,8). And the decoder is implemented as Dconv2(3,3,16)-Dconv2(3,2,32)-Dconv2(3,2,3). Except for the last layer, there are both batch normalized (BN) \cite{bn} and leaky ReLU activation \cite{leakyReLU} following a convolution layer. It is worth noting that we processed the MNIST and USPS datasets as RGB images. The domain discriminator $C$ is composed of linear layers, we define Linear($i$, $o$) to represent a linear layer with $i$-dimensional input and $o$-dimensional output. The domain discriminator $C$ is composed of three Linear layer Linear(72,128)-Linear(128,128)-Linear(128,1). Except for the final linear layer, each layer is followed by ReLU activation. We also use the dropout mechanism with a probability of 0.5. 
        
        We compare the proposed method with both of the traditional anomaly detection method and the deep learning-based anomaly detection methods. These baseline methods include iForest\cite {isolation}(unsupervised anomaly detection method), MemAE\cite {memae} (semi-supervised anomaly detection method). It is worth noting that we have also compare some variants of MemAE to show the importance of the proposed new components. We denote MemAE trained in the source domain as MemAE(directly) and denote MemAE trained in the source domain, then fine-tuned with the target domain as MemAE (fine-tune) MemAE(directly) and MemAE(fine-tune) are both compared to illustrate the necessity of weighted adversarial learning. The basis for judging anomalies in these two methods is different, for the tree-based method (such as iForest) and reconstruction-based methods (such as Mem
        AE, MemAE(directly), MemAE(fine-tune)), leaf node depth and reconstruction error are used to calculate anomaly scores. In all experiments, MemAE, MemAE(directly) and MemAE (fine-tune) share similar the structure settings, details can be found in \cite{memae}. Similarly, our method and all reconstruction-based methods use the same structure of encoder and decoder. AUC is used as a measure of performance, which is obtained by calculating the area under the Receiver Operation Characteristic(ROC). The table \ref{tab:digit} shows the average AUC value on the digital data set.
        
        As shown in the table \ref{tab:digit}, in addition to MemAE, the proposed method is better than other baselines on 8 transfer tasks out of 10 tasks. This is reasonable since our method only uses source normal samples and target unlabeled samples while MemAE directly uses target normal samples for training. The unsupervised anomaly detection method(IsolationForest) achieves a high AUC at a low anomaly rate, while the AUC decreases sharply as the anomaly rate increases. The performance of our method exceeds IsolationForest when the anomaly rate is greater than 0.25. In the two transfer tasks(MNIST$\rightarrow$USPS, 
        USPS$\rightarrow$MNIST), MemAE(directly) achieves the AUC of about 30 and 50 respectively. In fact, this indicates that the anomaly detection model trained in the source domain can not achieve an acceptable performance when applied directly to the target domain. Although the AUC on U$\rightarrow$M is about 20 higher than the AUC on M $\rightarrow$U, it is meaningless. On both tasks, the model can not detect anomalies effectively. The appearance of the AUC of 30 is caused by certain characteristics of handwritten digital images. MemAE(fine-tune) also achieves a relatively high AUC in the case of a low anomaly rate and a relatively low anomaly rate in the case of a high anomaly rate since high anomaly rate means that the target data will introduce more errors to model training. The performance of our method also decreases as the anomaly rate increases, however, the downward trend is not large since we reduce the impact of abnormal samples on the model training process by the weighting mechanism.
        
        The purpose of anomaly detection on video data is to identify anomalous events in the video. In this paper, the proposed method hopes to make use of the knowledge learned from the source video to detect anomalous events in the target video. We conduct experiments using two real-world video anomaly detection datasets, namely UCSD-ped1 \cite{mahadevan2010anomaly} and UCSD-ped2 \cite{mahadevan2010anomaly}, and construct two anomaly detection transfer tasks using these two datasets . The UCSD anomaly detection dataset is obtained from a camera fixed at a high position in the school, overlooking the sidewalk. The density of people in aisles varies from sparse to very crowded. In the normal setting, the video only contains pedestrians. Abnormal events are caused by the following two conditions: the circulation of non-sidewalk objects on the sidewalk, i.e. truck, bicycle, and unusual pedestrian movement patterns, i.e. the act of walking through the lawn, running. Ped1 contains the scene of people walking towards and away from the camera, including 34 training videos and 36 test videos. Ped 2 contains a scene where pedestrians move parallel to the camera plane, including 16 training videos and 12 test videos.
        
\begin{figure}
            \centering
            \includegraphics[width=.5\textwidth]{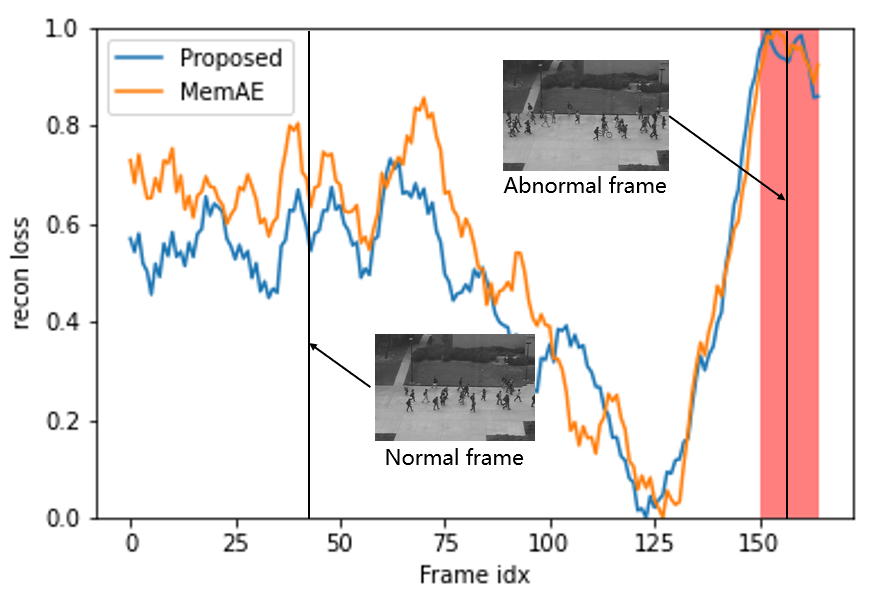}
            \caption{Reconstruction loss of the video frames obtain by our method and MemAE(fine-tune). When the anomalies appear, the reconstruction loss of our method increases larger than MemAE(fine-tune).}
            \label{fig:comrecon}
        \end{figure}

        Following the settings in \cite{memae}, for the purpose of making full use of the information contained in time dimension, 3D convolutional layers are used to build an antoencoder to extract the spatio-temporal features in the video. Therefore, the input of the network is a cuboid constructed by stacking 16 adjacent grayscale frames. The encoder contains the following structure: Conv3(3,(1,2,2),96)-Conv3(3,2,128)-Conv3(3,3,256)-Conv(3,2,256) and The decoder contains the following structure: Dconv3(3, 2,256)-Dconv3(3,2,128)-Dconv3(3,2,96)-Dconv3(3,(1,2,2),1), where Conv3 and Dconv3 represent 3D convolution layer and deconvolution layer, respectively. It is worth noting that a BN and a leaky ReLU activation follow each layer also except for the last layer similar to the structure used in image dataset. We set the input height and width as 256. In the inference phase, we obtain the normality score by reconstruction loss. The larger the reconstruction loss, the more likely the frame is to be an abnormal frame. From the Figure \ref{fig:comrecon}, we can find that the reconstruction loss immediately increases in our method when some abnormal events appear(bicycle in the street) in the video, and the increase in reconstruction loss between normal frames and abnormal frames is significantly greater than the method without knowledge transfer.
        
\begin{figure}
            \begin{subfigure}{.25\textwidth}
              \centering
              \includegraphics[width=\linewidth]{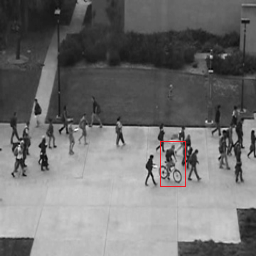}  
              \caption{Original frame}
            \end{subfigure}
            \begin{subfigure}{.25\textwidth}
              \centering
              \includegraphics[width=\linewidth]{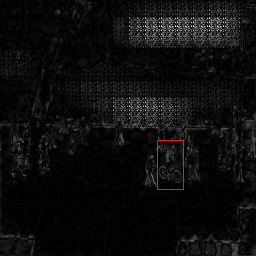}  
              \caption{MemAE}
            \end{subfigure}
            \begin{subfigure}{.25\textwidth}
              \centering
              \includegraphics[width=\linewidth]{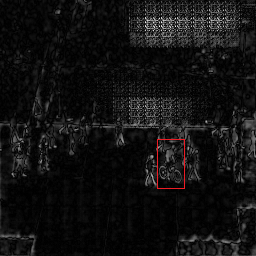}  
              \caption{Proposed}
            \end{subfigure}
            \caption{Reconstruction error of MemAE(fine-tune) and proposed method on an abnormal frame of UCSD-Ped2. Our method can significantly highlight the abnormal parts(in {\color{red}red} bounding box) in the scene.}
            \label{fig:videorecon}
        \end{figure}

        As shown in Table \ref{tab:video}, we compare the performance of our method with 5 state-of-the-art unsupervised anomaly detection methods. On both of the transfer tasks, our method consistently achieves the best performance. Sp + iForest \cite{isolation} uses the pre-trained Resnet50 as a feature extractor, and then cuts the obtained feature space. The algorithm stops when there is only one sample left in the splited feature space. Sp + iForest uses the depth of the leaf node as the anomaly detector. To have a fair comparison with the discriminative framework-based method, following the experimental setting in \cite{pang2020self}, our method is compared with its two variants: the first variant, namely Del Giorno et al. \cite{del2016discriminative} \#1 in Table \ref{tab:video}, uses ResNet-50 and PCA to extract features as the input to the discriminative framework; and the second variant, namely Del Giorno et al. \cite{del2016discriminative} \#2 in Table \ref{tab:video}, uses the features extracted from the last dense layer in the model of \cite{pang2020self}. Ionescu et al. \cite{tudor2017unmasking} and Liu et al. \cite{liu2018classifier} are both the unmasking framework, and the unmasking method is improved by a two-sample test method in \cite{liu2018classifier}. As can be seen from Table \ref{tab:video}, since the proposed method uses source normal samples to identify abnormal samples in target domain, compared with unsupervised methods, our method has achieved a significant performance improvement. Our method is also compared with other domain adaptation methods: MemAE(directly)(trained in source domain and tested in target domain directly) and MemAE(fine-tune)(trained in source domain and then fine-tuned with target samples). Importance weighted adversarial learning reduces the impact of abnormal samples in target domain in the training process, thus our method achieves the best performance among all domain adaptation methods. Semi-supervised method MemAE takes target normal samples as training samples directly, it is reasonable that it can achieve better performance than our method. However, the AUC of our method is only 2\% lower than the semi-supervised method, which is sufficient to demonstrate the effectiveness of our method. Figure\ref{fig:videorecon} visualizes an abnormal event on an abnormal frame in UCSD-Ped2. The reconstructed error map of our method highlights anomalous events (bicycles moving on the street). However, MemAE (fine tuning) can reconstruct the anomaly better than ours. This shows that our method can better identify anomalies.
        
        \begin{figure}
            \centering
            \includegraphics[width=0.5\textwidth]{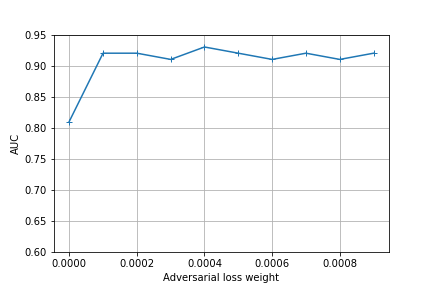}
            \caption{Robustness to the different adversarial loss weight $w_{adloss}$.}
            \label{fig:adlossweight}
        \end{figure}
        
        \subsection{Analysis Experiments}
            
            In this section, we use the handwritten digital dataset to perform some analysis experiments to illustrate the importance of the weighted adversarial learning.
            
            We visualize the reconstruction loss distributions of source samples and target samples in different training epochs. Figure \ref {fig:recondistribution} shows the reconstruction loss distributions of the task USPS $\rightarrow$ MNIST from 0 to 79 periods. Since importance weighted adversarial learning attempts to align the distributions of normal samples in the source and target domains, while keeping the distribution of abnormal samples in the target domain unchanged, the network has fully learned how to reconstruct normal samples in the source and target domains, instead of learning to reconstruct abnormal samples in the target domain. In this way, the reconstruction loss distributions of normal samples and abnormal samples in the target domain is pulled apart.
            
            In Figure \ref{fig:recon}, we visualize the image reconstruction results of different methods(MemAE(fine-tune), proposed method). Since the domain adaptation module in our method makes use of source samples to to help to reconstruct normal samples in target domain, given an abnormal input, the proposed method trained on '0' reconstructs an abnormal image, resulting in significant larger reconstruction error on the abnormal input than MemAE(fine-tune). The MemAE(fine-tune) do not have a weighted importance adversarial learning to help transfer knowledge from source domain, thus the anomalies in target domain will help MemAE(fine-tune) to learn how to reconstruct anomalies, resulting in the decrease of anomaly detection performance.
            
            \begin{table}[]
            \begin{tabular}{cc|cc}
            \hline
            \multicolumn{2}{c}{Method}                                                                                   & ped1$\rightarrow$ped2 & ped2$\rightarrow$ped1 \\ \hline
            \multicolumn{1}{c|}{\multirow{6}{*}{\rotatebox{90}{Un.}}} & Ionescu et al. \cite{tudor2017unmasking}      & 82.2                  & 68.4                  \\
            \multicolumn{1}{c|}{}                                                                 & Liu et al. \cite{liu2018classifier}          & 87.5                  & 69.0                  \\
            \multicolumn{1}{c|}{}                                                                 & Del Giorno et al. \cite{del2016discriminative} \#1 & 63.0                  & 50.3                  \\
            \multicolumn{1}{c|}{}                                                                 & Del Giorno et al. \cite{del2016discriminative} \#2 & 57.6                  & 59.6                  \\
            \multicolumn{1}{c|}{}                                                                 & Sp + iForest \cite{isolation}         & 67.5                  & 56.3                  \\
            \multicolumn{1}{c|}{}                                                                 & Guansong Pang et al. \cite{pang2020self} & 83.2                  & 71.7                  \\ \hline
            \multicolumn{1}{c|}{\multirow{3}{*}{\rotatebox{90}{Tran.}}}     & MemAE(directly) \cite{memae}      & 87.3                  & 75.0                  \\
            \multicolumn{1}{c|}{}                                                                 & MemAE(fine-tune) \cite{memae}     & 82.0                  & 73.0                  \\
            \multicolumn{1}{c|}{}                                                                 & Our method           & \textbf{91.9}         & \textbf{96.4}         \\ \hline
            \multicolumn{1}{c|}{\rotatebox{90}{Semi.}}               & MemAE \cite{memae}                & 94.0                  & -                    
            \end{tabular}
            \caption{AUC of different methods on video transfer tasks ped1$\rightarrow$ped2 and ped2$\rightarrow$ped1.}
            \label{tab:video}
            \end{table}
        
        In order to test whether the method is sensitive to adversarial loss weight $w_{adloss}$, we conducted the following experiments on the transfer task UCSD-Ped1 $\rightarrow$ UCSD-Ped2. The AUC of the anomaly detection model under different weight $w_{adloss}$ are recorded, and the results are shown in figure \ref{fig:lossweight}. From the figure, we can find that our proposed method is robust to weight $w_{adloss}$. Changes in weight will not cause significant changes in model performance.
        
        We study the inference complexity of this method on the video data set UCSD-Ped2 with the NVIDIA Tesla P100 16G graphics card. Our proposed method spends an average of 0.0613 seconds to determine whether a frame in the video is an abnormal frame, which spends a little more than the previous anomaly detection methods(for example, using\cite{liu2018future} using 0.04s, using\cite{luo2017revisit} using 0.02 seconds, \cite{tudor2017unmasking} uses 0.05 seconds.). In addition, compared with the baseline MemAE model(each frame \cite{memae} takes 0.0262 seconds), our importance-weighted adversarial learning module caused some additional calculations due to the domain classifier and weight module.
        
\section{Conclusion}
    This paper extends the adversarial nets-based domain adaptation to a new transfer anomaly detection scenario where the source domain only has normal data and the target domain has unlabeled normal/abnormal data. A weighting scheme based on reconstruction loss is proposed to effectively reduce the shift between source normal instances and target normal instances, and then detect the anomalous instances. The experimental results show that the proposed method outperforms previous unsupervised anomaly detection methods to a large degree and is comparable to the semi-supervised methods. For the future, we will further exploit the method to with focus on higher accuracy of identifying anomalies.

\section{Acknowledgement}
    We hereby give specical thanks to Alibaba Group for their contribution to this paper.

\clearpage
\bibliographystyle{ACM-Reference-Format}
\bibliography{ref}
\end{document}